\documentclass{article}

\usepackage{microtype}
\usepackage{graphicx}
\usepackage{subfigure}
\usepackage{booktabs} 

\usepackage{hyperref}

\usepackage[accepted]{icml2025}

\usepackage{amsmath}
\usepackage{amssymb}
\usepackage{mathtools}
\usepackage{amsthm}
\usepackage{multirow,tabularx}
\usepackage{makecell}
\usepackage{array}
\usepackage{ragged2e}
\newcolumntype{P}[1]{>{\RaggedRight\arraybackslash\hspace{0pt}}p{#1}}

\usepackage[capitalize,noabbrev]{cleveref}

\theoremstyle{plain}

\theoremstyle{definition}

\theoremstyle{remark}

\usepackage[textsize=tiny]{todonotes}

\icmltitlerunning{MOIRA: Multi-Omics Integration with Robustness to Absent modalities}

\begin{document}

\twocolumn[
\icmltitle{Robust Multi-Omics Integration from Incomplete Modalities \\ Significantly Improves Prediction of Alzheimer’s Disease}



\icmlsetsymbol{equal}{*}
\icmlsetsymbol{intern}{\dag}

\begin{icmlauthorlist}
\icmlauthor{Sungjoon Park}{equal,lgai}
\icmlauthor{Kyungwook Lee}{equal,lgai}
\icmlauthor{Soorin Yim}{lgai}
\icmlauthor{Doyeong Hwang}{lgai}
\icmlauthor{Dongyun Kim}{intern,lgai,snu}
\icmlauthor{Soonyoung Lee}{lgai}
\icmlauthor{Amy Dunn}{jax}
\icmlauthor{Daniel Gatti}{jax}
\icmlauthor{Elissa Chesler}{jax}
\icmlauthor{Kristen O'Connell}{jax}
\icmlauthor{Kiyoung Kim}{lgai}
\end{icmlauthorlist}

\icmlaffiliation{lgai}{LG AI Research, Seoul, Republic of Korea}
\icmlaffiliation{snu}{Department of Chemistry, Seoul National University, Seoul, Republic of Korea}
\icmlaffiliation{jax}{The Jackson Laboratory, Bar Harbor, Maine, USA}

\icmlcorrespondingauthor{Kiyoung Kim}{elgee.kim@lgresearch.ai}

\icmlkeywords{Machine Learning, ICML}

\vskip 0.3in
]



\printAffiliationsAndNotice{\icmlEqualContribution \textsuperscript{\dag}Work done during an internship at LG AI Research} 

\begin{abstract}
Multi-omics data capture complex biomolecular interactions and provide insights into metabolism and disease. However, missing modalities hinder integrative analysis across heterogeneous omics. To address this, we present MOIRA (Multi-Omics Integration with Robustness to Absent modalities), an early integration method enabling robust learning from incomplete omics data via representation alignment and adaptive aggregation. MOIRA leverages all samples, including those with missing modalities, by projecting each omics dataset onto a shared embedding space where a learnable weighting mechanism fuses them. Evaluated on the Religious Order Study and Memory and Aging Project (ROSMAP) dataset for Alzheimer’s Disease (AD), MOIRA outperformed existing approaches, and further ablation studies confirmed modality-wise contributions. Feature importance analysis revealed AD-related biomarkers consistent with prior literature, highlighting the biological relevance of our approach.
\end{abstract}

\section{Introduction}

Multi-omics integrates genome, transcriptome, proteome, metabolome, and epigenome to provide a comprehensive view of biological systems. This holistic approach is vital for capturing biological complexity \cite{skelly2019reference}, which is crucial for diseases like AD whose pathogenesis spans multiple layers. Recent advances in large-scale data generation and multi-modal learning have made such integration increasingly feasible and effective \cite{oh2021machine}.


However, challenges persist, particularly missing modalities, which are common due to diverse experimental protocols and limited sample availability \cite{flores2023missing, ballard2024deep}. This can lead to modality collapse during data integration without careful design \cite{javaloy2022mitigating}.

MOGONET \cite{wang2021mogonet} was the first to curate and release ROSMAP dataset \cite{perez2024rosmap} for AD prediction in machine learning field, establishing a benchmark widely adopted by subsequent studies. However, these studies were limited to omics modalities with relatively complete data, neglecting important sources such as proteomics \cite{bai2021proteomic}, and discarded samples with missing data even within the selected modalities.

In this study, we propose MOIRA, a novel method for phenotype prediction that accommodates missing modalities. We evaluate our approach on the ROSMAP dataset, which includes multi-omics data from AD patients and is characterized by a high degree of modality incompleteness. Our model effectively leverages incomplete multi-modal profiles and significantly outperforms prior methods on the AD prediction task.
Furthermore, utilizing multi-omics data facilitates the effective discovery of biomarkers \cite{jeong2023goat}. To this end, we identify relevant biomarkers using integrated gradients (IG) \cite{sundararajan2017axiomatic}, yielding results consistent with findings in the existing literature.

\begin{figure*}[t]
\begin{center}
\centerline{\includegraphics[width=2.09 \columnwidth]{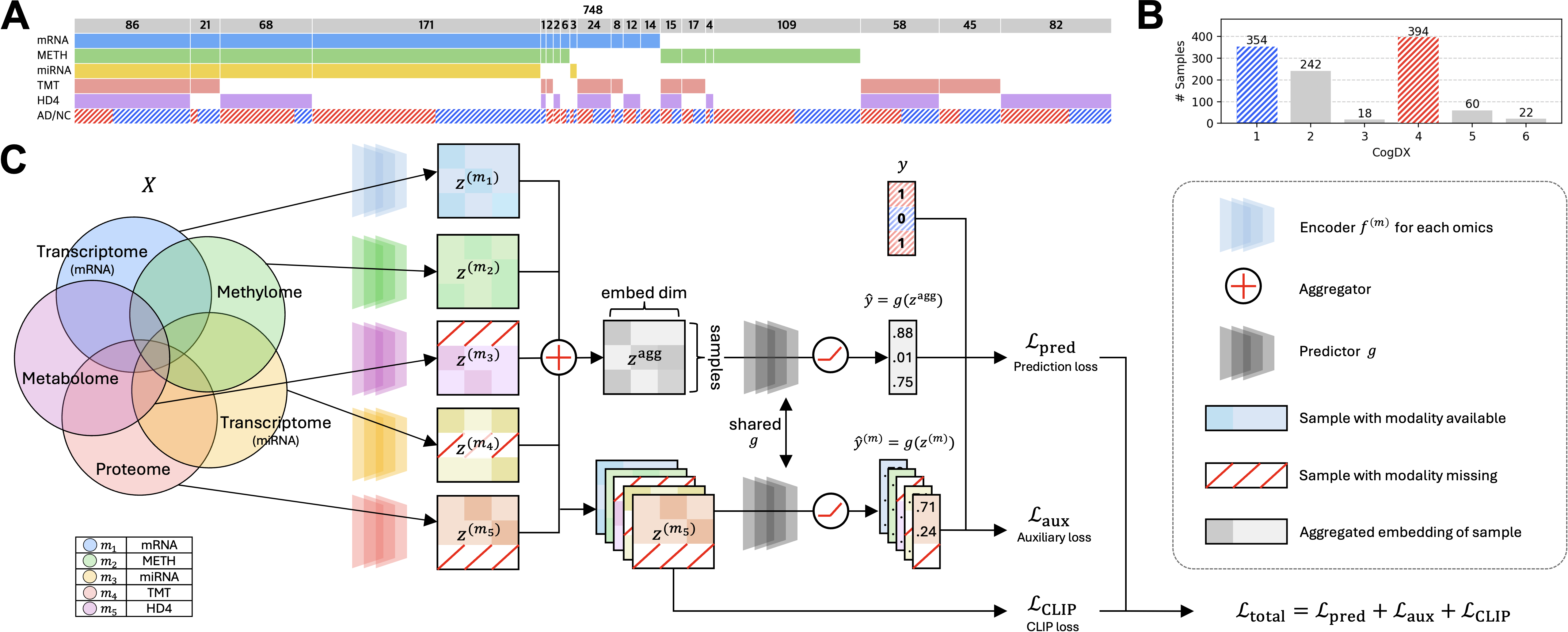}}
\caption{Overview of the ROSMAP dataset and the model architecture of MOIRA. \textbf{A)} Samples span multiple data modalities, with label:4 corresponding to AD and label:1 to normal controls (NC). Notably, only 86 (first column) samples contain complete data across all modalities. \textbf{B)} Distribution of samples by CogDX label, with majority categorized as label 1, 2, or 4. \textbf{C)} Model architecture. MOIRA consists of three phases: \textit{Encoders}, \textit{Aggregator}, and \textit{Predictor}. Each modality-specific \textit{Encoder} transforms heterogeneous input data into embedding vectors of a unified dimensionality. These embeddings are then integrated by the \textit{Aggregator} into a single representation. Finally, the \textit{Predictor} infers the class label based on this aggregated embedding. To address the challenge of missing modalities, the model is trained to align the embeddings from individual modalities, thereby reducing information loss and enhancing robustness.}
\label{overview}
\end{center}
\end{figure*}

\section{Materials and Methods}


\subsection{Data}

We used ROSMAP, a widely employed dataset in AD research. The dataset combines clinical, pathological and multi-omics measurements from aging individuals. These modalities—mRNA expression, DNA methylation (METH), microRNA (miRNA) expression, tandem mass tag (TMT) intensity and HD4 metabolite quantification—are all derived from post-mortem brain tissue, resulting in heterogeneous coverage and only 86 of 748 (11\%) samples having complete data across all five modalities (Figure~\ref{overview}A).

To train and evaluate our model, we focused on consensus cognitive diagnosis (CogDX), performing binary classification between label:4 (AD) and label:1 (no cognitive impairment), consistent with MOGONET \cite{wang2021mogonet} and related studies. These two labels dominate the class distribution in the ROSMAP dataset (Figure \ref{overview}B), making the binary classification task a practical simplification that effectively addresses the AD prediction problem.

\begin{table}[h]
\caption{Number of features and samples in the ROSMAP dataset.}
\label{features}
\begin{center}
\begin{small}
\begin{tabularx}{\linewidth}{Xccccc}
\toprule
& mRNA & METH & miRNA & TMT & HD4 \\
\midrule
\# features & 55,889 & 23,788 & 309 & 5,211 & 390 \\
\# samples & 630 & 740 & 521 & 400 & 514 \\
\bottomrule
\end{tabularx}
\end{small}
\label{n_features}
\end{center}
\end{table}

Given the variable feature dimensionality in each omics modality (Table~\ref{n_features}), we adopted MOGONET’s feature selection protocol. For each data type, we ranked features by their ANOVA F-scores on the training set and retained the top 200. This mitigates overfitting and prevents failure to capture important signals due to high dimensionality \cite{huang2022modality}, thereby improving model training.

\subsection{Model}
We introduce MOIRA for multi-omics integration that robustly handles missing modalities and mitigates modality collapse. The architecture comprises three main components: \textit{Encoders}, \textit{Aggregator}, and \textit{Predictor} (Figure~\ref{overview}C).

\paragraph{Encoders}
Each modality \( m \) is processed by a dedicated \textit{Encoder} \( f^{(m)} \), a two-layer MLP with LeakyReLU and dropout. Given input \( x^{(m)}_i \) for sample \( i \), it generates an embedding:
\[
z^{(m)}_i = f^{(m)}(x^{(m)}_i) \in \mathbb{R}^d
\]

\paragraph{Aggregator}
These modality-specific embeddings are then passed to the \textit{Aggregator}, which integrates them into a unified representation via a weighted sum.
\[
z^{\text{agg}}_i = \sum_{m \in \mathcal{M}_i} \alpha^{(m)}_i z^{(m)}_i, \quad \alpha^{(m)}_i = \frac{\exp(w^{(m)}_i)}{\sum_{n \in \mathcal{M}_i} \exp(w^{(n)}_i)}
\]

Here, \( \mathcal{M}_i \) is the set of observed modalities for sample \( i \); absent modalities have masked weights \( w^{(m)}_i = 0 \), ensuring that the softmax excludes them and reallocates their contribution among the present modalities.

\paragraph{Predictor}
The resulting aggregated embedding \( z^{\text{agg}}_i \) is then passed to the \textit{Predictor} \( g(\cdot) \), also a two-layer MLP with LeakyReLU and dropout, which produces the final output probability.
\[
\hat{y}_i = g(z^{\text{agg}}_i)
\]

The prediction loss is computed using cross-entropy.
\[
\mathcal{L}_{\text{pred}} = - \sum_i y_i \cdot \log(\hat{y}_i)
\]

To promote learning from individual modalities, we apply the same \textit{Predictor} to each modality-specific embedding and compute an auxiliary loss.
\[
\mathcal{L}_{\text{aux}} = - \sum_{m \in \mathcal{M}_i} \sum_i y_i \cdot \log(g(z^{(m)}_i))
\]

To prevent modality collapse and encourage alignment across modalities, we adopt a CLIP-style contrastive loss~\cite{radford2021learning}. For each modality pair \( (m, n) \), the directional contrastive loss is defined
\[
\mathcal{L}_{\text{CLIP}}^{(m \rightarrow n)} = - \frac{1}{N} \sum_{i=1}^{N} \log \frac{\exp(\text{sim}(z^{(m)}_i, z^{(n)}_i)/\tau)}{\sum_{j=1}^{N} \exp(\text{sim}(z^{(m)}_i, z^{(n)}_j)/\tau)},
\]

with cosine similarity \( \text{sim}(a, b) = \frac{a^\top b}{\|a\| \|b\|} \) and temperature \( \tau \). The full contrastive loss sums over all valid pairs.
\[
\mathcal{L}_{\text{CLIP}} = \sum_{m, n} \left( \mathcal{L}_{\text{CLIP}}^{(m \rightarrow n)} + \mathcal{L}_{\text{CLIP}}^{(n \rightarrow m)} \right)
\]

Finally, the total training objective combines all loss terms.
\[
\mathcal{L}_{\text{total}} = \mathcal{L}_{\text{pred}} + \mathcal{L}_{\text{aux}} + \mathcal{L}_{\text{CLIP}}
\]


\section{Results}

To evaluate the effectiveness of our method, we conducted experiments on the ROSMAP dataset, incorporating all samples with any available data modality. We used the same test split as MOGONET and related studies to ensure fair comparison. 
In all experiments, each encoder was paired with a decoder and pre-trained as part of an autoencoder until convergence using early stopping (patience = 30). Embedding dimensions were set to 300, with a dropout rate of 0.5. We used the Adam optimizer with a learning rate of 0.0001 and weight decay of 0.001, training for 200 epochs. All metrics were averaged over 30 independent runs.

\begin{figure}[h!]
\begin{center}
\centerline{\includegraphics[width=1.03\columnwidth]{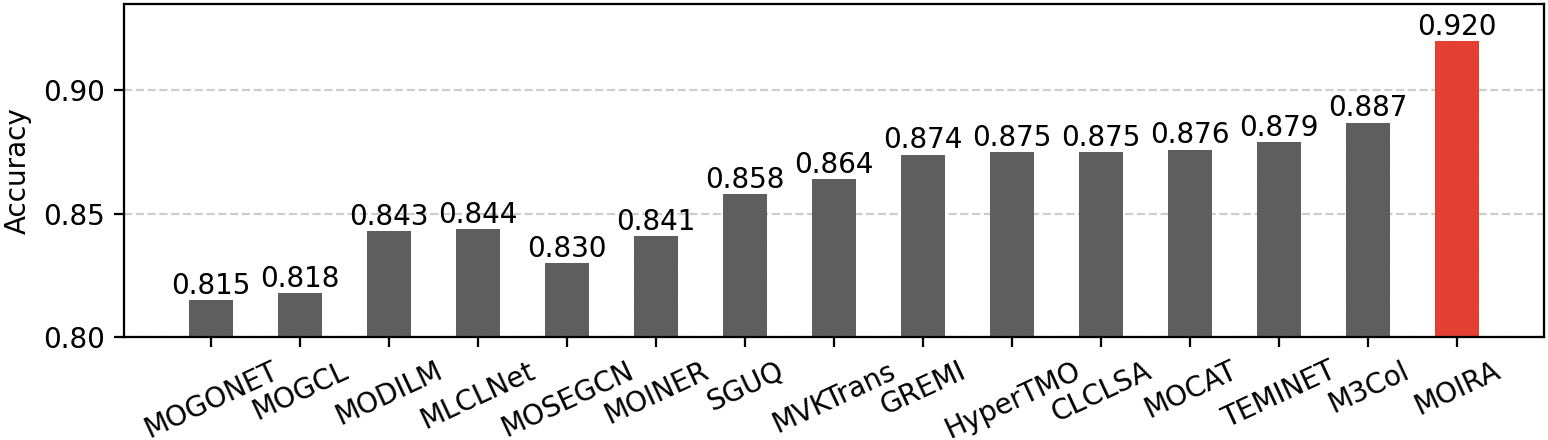}}
\caption{Performance comparison on AD prediction task.}
\label{performance-models}
\end{center}
\end{figure}

\paragraph{Outperforms SOTA methods in AD prediction}

Figure \ref{performance-models} compares the prediction accuracy of MOIRA with existing methods \cite{wang2021mogonet, rajadhyaksha2023graph, zhong2023modilm, zheng2023multi, wang2024semi, zhang2024moiner, tao2024sguq, cong2024mvktrans, liang2024gremi, wang2024hypertmo, zhao2024clclsa, yao2024mocat, luo2024teminet, kumarm3col} on the ROSMAP dataset. Our model achieved an accuracy of 0.920, significantly outperforming prior approaches. This gain arises from our ability to utilize incomplete data; unlike other models confined to 391 complete samples, it expands the usable set to 784 by incorporating partial modalities.


\begin{table}[h!]
\caption{Ablation studies using different combinations of data modalities and loss terms. Trimodal (i.e. mRNA+METH+miRNA) union ($\cup$) excludes only the additional modalities (TMT, HD4), while the intersection ($\cap$) further removes samples missing any of the remnant three. Minus (--) indicates silenced modality or loss.}
\label{ablation-studies}
\begin{center}
\begin{small}
\begin{tabularx}{\linewidth}{lcccc}
\toprule
& Accuracy & Precision & AUROC & AUPRC \\
\midrule
Full model & \textbf{0.920} & \textbf{0.972} & \textbf{0.922} & \textbf{0.914} \\
Trimodal ($\cup$) & 0.875 & 0.904 & 0.876 & 0.846 \\
Trimodal ($\cap$) & 0.867 & 0.874 & 0.867 & 0.826 \\
\midrule
--mRNA & 0.814 & 0.817 & 0.813 & 0.765 \\
--METH & 0.907 & 0.943 & 0.908 & 0.890 \\
--miRNA & 0.916 & 0.954 & 0.917 & 0.902 \\
--TMT & 0.894 & 0.937 & 0.896 & 0.876 \\
--HD4 & 0.916 & 0.956 & 0.917 & 0.903 \\
\midrule
--aux & \textbf{0.920} & 0.946 & 0.920 & 0.902 \\
--CLIP & 0.915 & 0.954 & 0.916 & 0.901 \\
--(aux+CLIP) & 0.896 & 0.938 & 0.898 & 0.878 \\
\bottomrule
\end{tabularx}
\end{small}
\end{center}
\end{table}

\paragraph{Ablation studies for data modality and loss terms}
To assess the contribution of each component in MOIRA to AD prediction, we conducted experiments across various input modalities and loss configurations. Table \ref{ablation-studies} summarizes evaluating under different conditions, including the omission of specific data modalities or loss terms.

We first evaluated performance using the three commonly used modalities—mRNA, METH, and miRNA. Trimodal ($\cup$) includes all samples with at least one of these modalities, while Trimodal ($\cap$) restricts to those with all three. We then systematically excluded individual modalities to assess their respective contributions to the AD prediction task. Lastly, we conducted ablation studies on the loss components.

Results show that using all five modalities—or any combination of four that includes mRNA—significantly outperforms the traditional three. Moreover, employing the union of samples yields better prediction than using only their intersection, highlighting the value of leveraging incomplete data rather than discarding partially observed samples. Finally, ablation of any individual loss term led to reduced performance when considering the overall metric profile.


\begin{table*}[h!]
\caption{Features extracted using IG scores and grouped by thematic relevance. In 100 repeated experiments, we selected the twelve most frequently occurring features from the top 10\% (i.e., the top 20 out of 200 features with the highest IG scores) for each modality. Corresponding references to the literature are indicated as numbers with brackets, and further listed in the final row.}
\label{igx}
\begin{center}
\begin{tiny}
\begin{tabularx}{\linewidth}{Xp{2.4cm}p{2.445cm}p{3cm}p{3cm}p{3.39cm}}
\toprule
Omics & Potential drug target & Prognostic biomarker & Important feature & Differential expression & Pathway/others \\
\midrule

mRNA &
SCD[1,2] &
KIF5A[3], HOPX[4] &
MEIS3[5,6], PPDPF[5], KIF5A[6-8], CSRP1[9-10], PLEKHB1[11], CDK2AP1[12], TAC3[13], SCD[8] &
MEIS3[14-17], PPDPF[14-16], KIF5A[16-18], PLEKHB1[15,17], HMGN2[15-17], QDPR[15-17,19], HOPX[15,17], CDK2AP1[15,17], PLEKHM2[20,21], TAC3[18,21] &
KIF5A[22], HMGN2[22,23] \\

\midrule
METH &
TMEM59[24,25], NGEF[26], PLEK[27] &
RORC[28], SNRPA[29] &
TMEM59[5,30,31], C10orf99[6,31], NGEF[6,30,31], RORC[31], SNRPA[5,6,31], KIAA1267[31], HSPA6[5,6], PLEK[5], LDHC[13], CHRLD2[32], HRASLS5[33] &
TMEM59[31], C10orf99[31], CHML[15,17,18], NGEF[31], RORC[31], SNRPA[31], KIAA1267[31], HSPA6[18], HRASLS5[34] &
TMEM59[35], C10orf99[36], SNRPA[37], KIAA1267[38,39], HSPA6[40], PLEK[41] \\

\midrule
miRNA &
miR-132[42,43] &
miR-132[44] miR-129-5p[44,45], miR-129-3p[44], let-7i[46], miR-125b[47-49], let-7g[46,50-52], miR-34a[49] &
miR-132[5,6,12,30,31], miR-129-5p[5,12,30,31], ebv-miR-BART8[13], miR-129-3p[5,12,30], miR-133b[5,13], miR-26a[10], let-7i[31] &
miR-132[31,53], miR-129-5p[31,53], let-7i[31], miR-125b[54], miR-34a[23] &
miR-132[55-58], miR-129-5p[57,58], miR-129-3p[57], miR-133b[57-59], miR-26a[56-58,60], let-7i[58], miR-29a[56-58], miR-9[57,58], miR-125b[55-58], let-7g[57,58], miR-34a[55-58] \\

\midrule
TMT &
NRN1[61], SLC38A2[62,63], MACROD1[64], GGT5[65] &
SMOC1[66], GFAP[67], CCK[68-69] &
SMOC1[70], SPOCK3[34,71], SPOCK2[71] &
GFAP[15,17,72,73], RAB27B[15,18], NRN1[15,18], SLC38A2[14,18,72], SPOCK3[74,75], GGT5[15] &
GFAP[23], SPOCK3[76,77], SPOCK2[78] \\

\midrule
HD4 &
Alpha-GPC[79], homocarnosine[80], threonate[81,82], myo-inositol[83,84], caprate (10:0)[85,86] &
pipecolate[87], N-Acetyl-GABA[88], myo-inositol[89], N6-methyllysine[90,91], dimethylarginine[92] &
pipecolate[93] &
carboxyethyl-GABA[94,95], X - 24035[94], dimethylarginine[96] &
Alpha-GPC[97], homocarnosine[98], myo-inositol[99], dimethylarginine[100,101] \\

\midrule
\multicolumn{6}{p{0.979 \linewidth}}{
*
[1] \citet{hamilton2022stearoyl}
[2] \citet{loix2024stearoyl}
[3] \citet{hares2019kif5a}
[4] \citet{liu2023early}
[5] \citet{liang2024gremi}
[6] \citet{luo2024teminet}
[7] \citet{wang2021mogonet}
[8] \citet{wang2024hypertmo}
[9] \citet{kong2009independent}
[10] \citet{briscik2024supervised}
[11] \citet{graham2025human}
[12] \citet{zhang2024moiner}
[13] \citet{wang2024semi}
[14] \citet{wang2023gene}
[15] \citet{li2021differentially}
[16] \citet{mccorkindale2022key}
[17] \citet{aguzzoli2025systematic}
[18] \citet{vastrad2021bioinformatics}
[19] \citet{rahimzadeh2024gene}
[20] \citet{liu2023early}
[21] \citet{tian2022identification}
[22] \citet{millecamps2013axonal}
[23] \citet{mathys2024single}
[23] \citet{schipper2007microrna}
[24] \citet{meng2020tmem59}
[25] \citet{liu2020tmem59}
[26] \citet{hudgins2024systems}
[27] \citet{dai2022hypothalamic}
[28] \citet{huang2025transcriptomic}
[29] \citet{jiang2018signaling}
[30] \citet{cong2024mvktrans}
[31] \citet{yao2024mocat}
[32] \citet{huang2022identification}
[33] \citet{zheng2024identifying}
[34] \citet{shu2022identification}
[35] \citet{ullrich2010novel}
[36] \citet{lee2011identification}
[37] \citet{hsieh2019tau}
[38] \citet{poorkaj2001genomic}
[39] \citet{prasad2019concise}
[40] \citet{wu2021analysis}
[41] \citet{samadian2021eminent}
[42] \citet{zhang2021alzheimer}
[43] \citet{walgrave2021restoring}
[44] \citet{nagaraj2024downregulation}
[45] \citet{han2024mir}
[46] \citet{derkow2018distinct}
[47] \citet{hong2017identification}
[48] \citet{yashooa2022mir}
[49] \citet{swarbrick2019systematic}
[50] \citet{poursaei2022evaluation}
[51] \citet{kafshdooz2023hsa}
[52] \citet{kumar2013circulating}
[53] \citet{noronha2022differentially}
[54] \citet{mckeever2018microrna}
[55] \citet{li2024micrornas}
[56] \citet{liu2022micrornas}
[57] \citet{kumar2016circulating}
[58] \citet{sun2021role}
[59] \citet{yang2019mir}
[60] \citet{xie2022identification}
[61] \citet{hurst2023integrated}
[62] \citet{li2022establishing}
[63] \citet{patel2019transcriptomic}
[64] \citet{carlyle2021synaptic}
[65] \citet{zhang2025gamma}
[66] \citet{balcomb2024smoc1}
[67] \citet{kim2023gfap}
[68] \citet{plagman2019cholecystokinin}
[69] \citet{zhang2023cholecystokinin}
[70] \citet{roberts2023unbiased}
[71] \citet{oveisgharan2024proteins}
[72] \citet{wang2021cell}
[73] \citet{jing2021comprehensive}
[74] \citet{ma2020fusiform}
[75] \citet{levites2023abeta}
[76] \citet{pan2020transcriptomic}
[77] \citet{wojtas2024proteomic}
[78] \citet{grupe2006scan}
[79] \citet{lee2017late}
[80] \citet{hipkiss2007could}
[81] \citet{liao2024magnesium}
[82] \citet{kim2020neuroprotective}
[83] \citet{ali2022brain}
[84] \citet{barak1996inositol}
[85] \citet{shekhar2023potential}
[86] \citet{fan2023association}
[87] \citet{gonzalez2015metabolite}
[88] \citet{wang2021uplc}
[89] \citet{voevodskaya2019brain}
[90] \citet{wang2021genome}
[91] \citet{panyard2021cerebrospinal}
[92] \citet{choi2020asymmetric}
[93] \citet{hammond2020metabolite}
[94] \citet{batra2023landscape}
[95] \citet{borghys2024middle}
[96] \citet{zinellu2023circulating}
[97] \citet{miatto1986vitro31p}
[98] \citet{balion2007brain}
[99] \citet{miller1993alzheimer}
[100] \citet{selley2003increased}
[101] \citet{popp2012p4}
} \\
\bottomrule

\end{tabularx}
\end{tiny}
\end{center}
\end{table*}

\paragraph{Features related to Alzheimer's Disease extracted}

To biologically validate our model, we identified the top features contributing to phenotype prediction using IG. For each modality, we selected twelve features with the highest IG values (representing top 16\%), and repeated the analysis 100 times with different random seeds to ensure robustness. Table \ref{igx} lists the selected features, their known or suspected relevance to AD, and corresponding literature. mRNA feature names were mapped to gene symbols using pyensembl, while METH features were matched to genes via CpG-to-gene mapping from the HumanMethylation450 BeadChip annotation.

Among the identified features, only LAGE3 and TTC33 from the TMT modality, and N-methylpipecolate from HD4, did not have any evidence for linkage to AD. For the remaining 57 features, we found at least one supporting source, classified as a potential drug target, prognostic biomarker, differentially expressed gene, or key feature reported in prior machine learning studies. We also noted indirect evidence of some features, such as AD-associated pathways. These results demonstrate that MOIRA is capable of feature attribution in datasets with incomplete modalities.

\section{Conclusion}

In this paper, we proposed MOIRA, a novel method to address the challenge of incomplete multi-omics data. Prior studies using ROSMAP focused only on samples with complete mRNA, DNA methylation, and miRNA data. In contrast, our model's ability to handle missing modalities allowed us to incorporate TMT and HD4 data, effectively doubling the training set size. As a result, MOIRA outperforms state-of-the-art methods in AD prediction on ROSMAP. Moreover, IG analysis reveals biologically meaningful features that are consistent with existing literatures on AD.

While our approach maximizes data utilization, there still is room for improvement. We did not incorporate the genomic dataset, due to its size and complexity. And although we address missingness at the modality level, feature-level absence of data samples remains unhandled. In future work, we aim to pre-train encoders using masked autoencoders to address feature-level missingness and leverage foundation models to integrate the genomic data. While this study focuses on multi-omics phenotype prediction, our scheme is domain-agnostic and can also be applied broadly to multi-modal learning scenarios where missing data are prevalent.

\section*{Acknowledgements}

This work was supported by LG AI Research and in part by the National Institutes of Health (NIH) grant
RF1AG059778, and the Alzheimer’s Association Research Fellowship AARF-18-565506. The results published here are in whole or in part based on data obtained from the AD Knowledge Portal (https://adknowledgeportal.org).
Study data were provided by the Rush Alzheimer’s Disease Center, Rush University Medical Center, Chicago. Data collection was supported through funding by National Institute on Aging (NIA) grants P30AG010161, R01AG015819, R01AG017917, R01AG030146, R01AG036836, U01AG046161, and the Illinois Department of Public Health (IDPH). Additional phenotypic data can be requested at www.radc.rush.edu.

\paragraph{RNA-seq bulk brain.}
Annie J. Lee, Yiyi Ma, Lei Yu, Robert J. Dawe, Cristin McCabe, Konstantinos Arfanakis, Richard Mayeux, David A. Bennett, Hans-Ulrich Klein, and Philip L. De Jager. Multi-region brain transcriptomes uncover two subtypes of aging individuals with differences in AD risk and the impact of APOE$\epsilon$4. bioRxiv 2021

\paragraph{Array expression.}
We thank the patients and their families for their selfless donation to further understanding AD. This project was supported by funding from the NIA (R01AG034504, R01AG041232). Many data and biomaterials were collected from several NIA and National Alzheimer’s Coordinating Center (NACC) funded sites (U01AG016976). Amanda J. Myers, PhD (University of Miami, Department of Psychiatry) prepared the series. The directors, pathologist and technicians involved include: Rush University Medical Center, Rush Alzheimer's Disease Center (P30AG010161): David A. Bennett, MD, Julie A. Schneider, MD, MS, Karen Skish, MS, PA(ASCP), MT, Wayne T Longman. The Rush portion of this study was supported by NIH grants P30AG010161, R01AG015819, R01AG017917, R01AG036042, R01AG036836, U01AG\allowbreak046152, R01AG034374, R01NS078009, U18NS082140, R01AG042210, R01AG039478, and the IDPH. Quality control checks and preparation of the gene expression data was provided by the National Institute on Aging Alzheimer’s Disease Data Storage Site (NIAGADS) at the University of Pennsylvania (U24AG041689).

\paragraph{TMT proteomics.}
Study data were provided through the Accelerating Medicine Partnership for AD (U01AG046161, U01AG061357) based on samples provided by the Rush Alzheimer’s Disease Center, Rush University Medical Center, Chicago. Data collection was supported through funding by NIA grants P30AG010161, R01AG015819, R01AG017917, R01AG030146, R01AG036836, U01AG\allowbreak032984, U01AG046152, the IDPH, and the Translational Genomics Research Institute.

\paragraph{Metabolomics.}
Metabolomics data is provided by the Alzheimer’s Disease Metabolomics Consortium (ADMC) and funded wholly or in part by the following grants and supplements thereto: NIA R01AG046171, RF1AG051550, RF1AG057452, R01AG059093, RF1AG\allowbreak058942, U01AG061359, U19AG063744 and FNIH: \#DAOU16AMPA awarded to Dr. Kaddurah-Daouk at Duke University in partnership with a large number of academic institutions. As such, the investigators within the ADMC, not listed specifically in this publication’s author’s list, provided data along with its pre-processing and prepared it for analysis, but did not participate in analysis or writing of this manuscript. A complete listing of ADMC investigators can be found at: https://sites.duke.edu/adnimetab/team/. The Metabolon datasets were generated at Metabolon and pre-processed by the ADMC.

\section*{Impact Statement}
This paper presents work whose goal is to advance the field of Machine Learning. There are many potential societal consequences of our work, none which we feel must be specifically highlighted here.



\bibliography{moira}
\bibliographystyle{icml2025}

\end{document}